\documentclass[10pt,twocolumn,letterpaper]{article}

\usepackage{cvpr} 
\usepackage{times}
\usepackage{epsfig}
\usepackage{graphicx}
\usepackage{amsmath}
\usepackage{amssymb}
\usepackage{booktabs}
\usepackage{multirow}
\usepackage{algorithm}
\usepackage{algorithmic}
\usepackage{hyperref}

\begin{document}

\title{Diff-IrisPAD: Modeling Bona Fide Iris Distributions for Unsupervised Presentation Attack Detection}

\author{
Geetanjali Sharma\textsuperscript{1}, 
Harsh Yadav\textsuperscript{1, *}, 
Tanmay Deshpande\textsuperscript{1, *}, 
Shubham Ashwani\textsuperscript{1}, \\
Aditya Nigam\textsuperscript{1}, 
Raghavendra Ramachandra\textsuperscript{2} \\
\textsuperscript{1}Indian Institute of Technology Mandi, Himachal Pradesh, India \\
\textsuperscript{2}SAFE Center, Norwegian University of Science and Technology (NTNU), Gjøvik, Norway \\
}
\maketitle
\footnotetext[1]{* These authors contributed equally to this work.}
\footnotetext{https://github.com/GeetanjaliGTZ/DiffIrisPAD}

\thispagestyle{empty}
\begin{abstract}
Iris recognition systems are vulnerable to presentation attacks (PAs), including textured contact lenses, print replays, and synthetic irises. Existing PAD methods rely on supervised learning with attack data, limiting generalization to unseen attacks and new sensors.
We propose \textbf{Diff-IrisPAD}, an unsupervised framework based on Brownian Bridge Diffusion that models bona fide iris distributions without using attack samples. The method reconstructs clean iris textures from degraded inputs and detects attacks via reconstruction inconsistency using ViT scoring and pixel-level measures.
Experiments on a large-scale multi-sensor dataset (125K+ images, NIR and VIS, 8 attack types) show that Diff-IrisPAD achieves an overall ACER of \textbf{30.52}, significantly outperforming diffusion-based baselines while demonstrating strong generalization to unseen attacks. These results highlight the effectiveness of diffusion-based bona fide modeling for robust and scalable iris PAD.
\end{abstract}

\section{Introduction}

Iris recognition is one of the most reliable biometric technologies due to the distinctive and stable texture patterns of the human iris. It is widely deployed in applications such as border control, national identity systems, and mobile authentication~\cite{zhang2016btas,iab_iris}. Despite its success, iris recognition systems remain vulnerable to \textit{presentation attacks} (PAs), where adversaries attempt to spoof the sensor using printed images, textured contact lenses, electronic displays, or synthetic irises~\cite{gupta2014iris,yadav2014unraveling}.

To mitigate these threats, iris Presentation Attack Detection (PAD) has been extensively studied. Early approaches relied on handcrafted features, including texture descriptors for detecting contact lenses~\cite{doyle2015robust} and multi-feature fusion strategies~\cite{kohli2016detecting}, as well as imaging-based cues such as light field characteristics~\cite{raghavendra2014presentation}. Benchmark efforts like LivDet-Iris~\cite{yambay2017livdet} have further standardized evaluation and accelerated progress in this field. More recently, deep learning-based methods have significantly improved PAD performance by learning discriminative representations directly from data.
However, most existing PAD approaches follow a supervised learning paradigm that requires both bona fide and attack samples during training. This assumption is restrictive, as new and unseen attack types continuously emerge. Consequently, supervised models often overfit to known attacks and exhibit significant performance degradation under cross-sensor and unseen attack scenarios~\cite{yambay2017livdet,crossdevice2025ipad,wang2025attention}. This challenge is further exacerbated by the increasing realism of attacks, including textured contact lenses, high-quality display attacks, post-mortem samples, and synthetic irises generated by modern generative models. These limitations highlight the need for approaches that can generalize beyond known attack categories.
Recent works attempt to address these challenges using open-set PAD and foundation model-based approaches~\cite{tapia2025foundation,dowling2026viser}. While promising, these methods still rely on labeled attack data and do not fully solve the problem of generalization to unseen attacks. This motivates the need for fundamentally different approaches that model only bona fide data.

In this work, we propose \textbf{Diff-IrisPAD}, an unsupervised framework for iris PAD based on Brownian Bridge Diffusion. Unlike conventional diffusion models that learn noise-to-data mappings, BBDM enables a direct transformation between degraded and clean images, resulting in more efficient and stable reconstruction. 
The proposed method models the distribution of bona fide iris images by learning to reconstruct clean iris textures from synthetically degraded inputs. During inference, both bona fide and attack samples are processed through the same reconstruction pipeline. Since the model is trained exclusively on bona fide data, it reconstructs genuine samples accurately while failing to reproduce attack-specific artifacts. This discrepancy is quantified using perceptual similarity metrics such as LPIPS~\cite{zhang2018lpips}.
The proposed framework is inspired by diffusion-based anomaly detection~\cite{wyatt2022anoddpm}, but is specifically designed for iris PAD, addressing the challenges of fine-grained texture modeling and cross-attack generalization.
\noindent\textbf{Contributions.} The main contributions of this work are:
\begin{itemize}
\item \textbf{Diff-IrisPAD}: An unsupervised iris presentation attack detection framework based on BBDM, which models the distribution of bona fide iris images without requiring attack samples.

\item Reconstruction-based \textbf{ PAD formulation} in latent space, where attacks are identified via reconstruction inconsistency using ViT metrics.

\item A \textbf{latent diffusion architecture} integrating a frozen VQGAN with a trainable BBDM, enabling efficient and stable learning of fine-grained iris texture representations.

\item An \textbf{attack-agnostic learning paradigm} that improves robustness to unseen presentation attacks by relying solely on bona fide data.

\item Extensive evaluation on a multi-sensor iris PAD dataset, demonstrating competitive performance and improved generalization compared to diffusion-based baselines.
\end{itemize}

\section{Related work}
Deep learning has significantly advanced iris Presentation Attack Detection (PAD), with early strong baselines shaping later research. A cascade-based CNN framework was introduced by \cite{tapia2022cascade}, where multiple specialised networks focus on different attack types and their outputs are fused hierarchically. While effective, this approach highlighted the need for richer feature representations, which motivated \cite{agarwal2022enhanced} to propose a dual-stream contraction–expansion CNN that jointly processes raw and edge-enhanced iris images to capture complementary cues and improve generalisation. Moving beyond single-task learning, \cite{dhar2022eyepad++} presented EyePAD++, a teacher–student framework that jointly learns iris recognition and PAD, showing that identity features can enhance spoof detection. Building on the need for robustness in real-world settings, \cite{agarwal2022generalized} focused on generalised contact lens detection, addressing unseen lens types and demonstrating the importance of handling evolving attack patterns.

Despite these advances, a key challenge that emerged is domain shift, where models fail under unseen sensors or environments. \cite{czajka2018presentation} highlighted this limitation, showing that models performing well in controlled settings degrade significantly in cross-domain scenarios. To better understand this, \cite{fang2023crossspectrum} analysed cross-spectrum PAD across NIR and visible domains, revealing large performance gaps when models are evaluated across imaging conditions. Extending this, \cite{fang2022augmentation} studied data augmentation strategies and found that naïve combinations can harm performance, emphasising the need for structured augmentation. Addressing this issue, \cite{li2023dynamic} proposed a Single Domain Dynamic Generalisation framework that simulates domain shifts using only one dataset, while \cite{pal2024adversarial} further improved robustness using adversarial augmentation to generate challenging training samples without requiring target-domain data.

These limitations naturally led to the exploration of open-set PAD, where attack types during testing are unseen. \cite{boyd2023openset} demonstrated that most existing methods fail in such settings, motivating a shift toward anomaly detection and one-class learning approaches that do not rely on predefined attack categories.
At the same time, researchers improved feature learning using attention and fusion strategies. \cite{swarup2022attention} introduced attention mechanisms to help models focus on important iris regions, while \cite{deng2025attention} extended this idea with a multilevel fusion framework that separates global and local attack cues, achieving strong performance in recent benchmarks.
Another important direction considers visible-spectrum and periocular-based PAD, as most earlier works focus on NIR images. \cite{valenzuela2024periocular} explored PAD in RGB images and showed that performance drops significantly across datasets, reinforcing the generalisation problem. To mitigate data scarcity, \cite{maureira2022synthetic} demonstrated that synthetic periocular data can partially improve training, especially when real attack samples are limited. 
In parallel, privacy concerns have driven research into synthetic and privacy-preserving training. \cite{purnapatra2024privacy} showed that PAD models can be trained entirely on synthetic iris data without using real subjects, addressing ethical challenges. This aligns with benchmark efforts such as LivDet-Iris 2023\cite{tinsley2023livdet}, which introduced GAN-generated attacks into evaluation protocols. Further advancing this direction, \cite{bhuiyan2024forensic} proposed forensic iris synthesis to generate realistic attack samples and analyse model weaknesses.

More recently, foundation models and large vision models have started influencing iris PAD. \cite{tapia2025foundation} showed that features from pretrained models such as DINOv2 can outperform traditional approaches when data is limited, while\cite{farmanifard2024chatgpt} explored zero-shot iris analysis using large language-vision models, opening new research directions. Complementing these developments, \cite{yin2023review} provided a comprehensive survey highlighting the growing role of transformer-based models in iris-related tasks.
The LivDet-Iris competitions remain the primary benchmark for evaluating PAD systems. The 2023 edition\cite{tinsley2023livdet} introduced diverse attack types including synthetic irises, while the 2025 edition \cite{livdet2025} further expanded the evaluation to include unknown attacks, dataset aging, and morphing-based attacks, reinforcing the need for robust and generalisable PAD solutions.

\section{Methodology}
Diff-IrisPAD is an unsupervised iris PAD framework based on Brownian Bridge Diffusion that models bona fide iris distributions and detects attacks via reconstruction inconsistency.

\subsection{Overview of Diff-IrisPAD}
\label{sec:overview}

The proposed Diff-IrisPAD framework follows a reconstruction-based learning paradigm, as illustrated in Figure.~\ref{fig:Diff-iris-PAD}. The model is trained exclusively on bona fide iris images, where each clean sample is paired with a synthetically degraded version. A diffusion model learns to reconstruct clean iris textures from these degraded inputs.
During inference, both bona fide and attack samples are processed through the same reconstruction pipeline. Since the model captures only the distribution of genuine iris images, it reconstructs bona fide samples accurately while failing to reproduce artifacts present in attack samples. This difference in reconstruction quality is used to distinguish between bona fide and presentation attacks. The overall framework consists of three stages: (i) diffusion-based reconstruction, (ii) computation of reconstruction and trajectory-based scores, and (iii) threshold-based decision making.

\subsection{Brownian Bridge Diffusion Model}
\label{sec:bbdm}
Standard denoising diffusion models (DDPMs) learn a noise-to-data mapping, which is inefficient for paired image tasks. In contrast, Brownian Bridge Diffusion Models (BBDM)~\cite{li2023bbdm} learn a direct mapping between source and target images using a stochastic bridge process.
We adopt the BBDM~\cite{li2023bbdm} to learn a mapping between degraded and clean iris images. Unlike standard DDPMs~\cite{ho2020ddpm}, which model a noise-to-data trajectory, BBDM directly learns a stochastic bridge between paired images, enabling more efficient reconstruction.

Given a degraded input $\mathbf{x}_0$ and its corresponding clean target $\mathbf{y}$, the forward process is defined as:
\begin{equation}
q(\mathbf{x}_t \mid \mathbf{x}_0, \mathbf{y}) =
\mathcal{N} \big( (1-m_t)\mathbf{x}_0 + m_t\mathbf{y}, \ \delta_t \mathbf{I} \big),
\end{equation}
where $m_t$ controls interpolation and $\delta_t$ defines the variance schedule.
The model is trained to predict the reverse process using a UNet architecture, minimizing the L1 loss between predicted and true intermediate states. To improve efficiency, we use the latent variant (LBBDM-f4), which encodes images into a compressed latent space using a pretrained VQGAN~\cite{esser2021vqgan}, significantly reducing computational cost. 
This formulation allows the model to learn fine-grained iris texture distributions while maintaining stable training dynamics.




\subsection{Unsupervised Iris PAD Formulation}
\label{sec:unsupervised_pad}
Unlike conventional PAD methods that require both bona fide and attack samples, our approach formulates iris PAD as an unsupervised reconstruction problem.
During training, only bona fide iris images are used. For each image $\mathbf{y}$, a degraded version $\mathbf{x}_0$ is generated using a degradation function:
\begin{equation}
\mathbf{x}_0 = \mathcal{C}(\mathbf{y}),
\end{equation}
where $\mathcal{C}(\cdot)$ includes noise injection, blurring, and resolution degradation. The degradation function applies only structure-preserving corruptions rather than simulating presentation attacks. This allows Diff-IrisPAD to learn the bona fide iris manifold from paired clean and degraded images, treating attack samples as out-of-distribution during inference.
The model is trained to reconstruct $\mathbf{y}$ from $\mathbf{x}_0$, thereby learning the distribution of genuine iris textures.
At inference, both bona fide and attack samples are processed in the same manner. Since the model has only learned bona fide characteristics, it generalizes poorly to attack samples, resulting in higher reconstruction errors. This enables detection of unseen presentation attacks without requiring attack data during training. This effectively treats presentation attacks as out-of-distribution samples relative to the learned bona fide manifold.

\begin{figure*}[ht]
\centering
\includegraphics[width=0.8\linewidth]{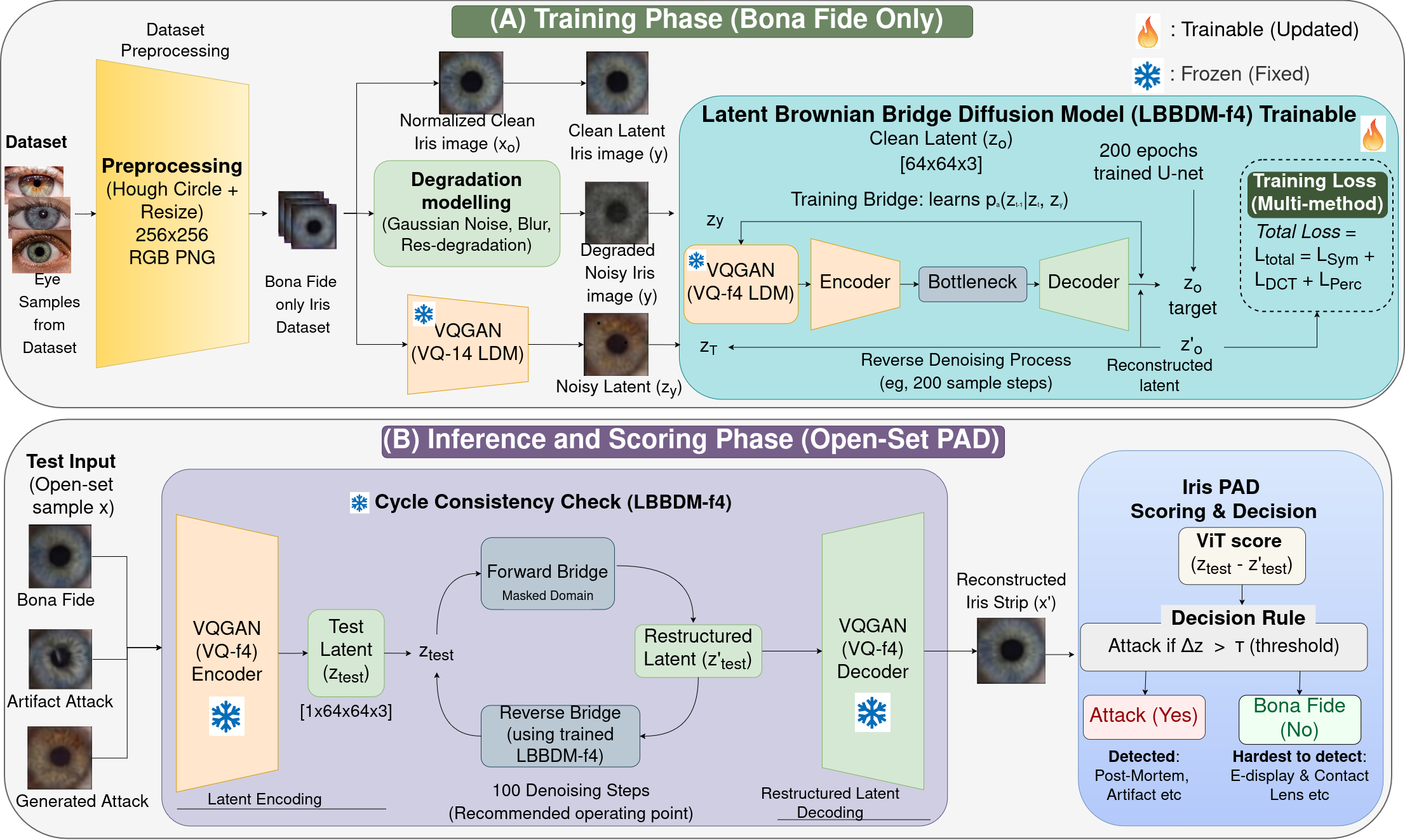}
\caption{
\textbf{Diff-IrisPAD framework.}
(A) Training: a latent BBDM (LBBDM-f4) learns bona fide iris distributions by reconstructing clean images from degraded inputs using a frozen VQGAN and trainable UNet. 
(B) Inference: reconstruction inconsistency ViT is used for threshold-based open-set attack detection without using attack data.
}
\label{fig:Diff-iris-PAD}
\end{figure*}


\subsection{Diff-IrisPAD Architecture}

We adopt the LBBDM-f4 framework~\cite{li2023bbdm}, which operates in a compressed latent space obtained using a frozen VQ-f4 encoder~\cite{esser2021vqgan}. This latent representation reduces spatial redundancy while preserving fine-grained iris texture essential for PAD. The encoder uses an embedding dimension of 3, a codebook size of 8,192, and performs $4\times$ spatial downsampling.

The denoising process is modeled using a UNet backbone with \texttt{model\_channels}$=128$ and channel multipliers $[1,2,3,4]$. Each resolution stage contains two residual blocks, and multi-head self-attention (8 heads, 64 channels each) is applied at resolutions $\{32,16,8\}$ to capture both local texture variations and global structural cues. Notably, no explicit conditioning signal is injected into the network; instead, the clean iris representation $\mathbf{y}$ initializes the reverse diffusion process through its latent encoding, enabling implicit conditioning.

The diffusion process is defined over $T=1{,}000$ timesteps with a linear $m_t$ schedule and \texttt{max\_var}$=1.0$. Training is guided by an $\ell_1$ objective on the predicted bridge increment (Eq.~\ref{eq:bbdm_loss}), which stabilizes reconstruction and improves convergence. During inference, we employ DDIM-style sampling with \texttt{eta}$=1.0$ and a linear skip schedule, allowing efficient reconstruction with reduced sampling steps.
\begin{equation}
  \mathcal{L}_{\mathrm{BBDM}}
  = \mathbb{E}_{t,\mathbf{x}_0,\mathbf{y},\boldsymbol{\epsilon}}\!\left[
      \bigl\|
        m_t(\mathbf{y} - \mathbf{x}_0)
        + \sqrt{\delta_t}\,\boldsymbol{\epsilon}
        - \mu_\theta(\mathbf{x}_t, t)
      \bigr\|_1
    \right],
  \label{eq:bbdm_loss}
\end{equation}

Optimisation is performed using Adam with learning rate $1{\times}10^{-4}$, $\beta_1=0.9$, and zero weight decay. A \texttt{ReduceLROnPlateau} scheduler adaptively reduces the learning rate based on validation performance (factor 0.5, patience 3,000 steps, cooldown 3,000 steps, $\text{lr}_{\min}=5{\times}10^{-7}$). Gradient accumulation over 4 micro-steps results in an effective batch size of 64. To improve stability, exponential moving average (EMA) with decay 0.995 is applied, updated every 8 steps and activated after 30,000 warm-up steps.
Training is conducted on a single NVIDIA RTX A5000 with seed 42. The model is trained for 200 epochs (${\approx}410{,}000$ steps), with early convergence observed at epoch 70 (${\approx}$55 h 45 min). 

\subsection{Training Strategy(Diff-IrisPAD)}
\label{sec:training}

The model is trained exclusively on bona fide iris images using paired data $(\mathbf{x}_0, \mathbf{y})$, where $\mathbf{y}$ denotes a clean iris image and $\mathbf{x}_0 = \mathcal{C}(\mathbf{y})$ is its synthetically degraded counterpart. The degradation function $\mathcal{C}(\cdot)$ applies a sequence of perturbations, including Gaussian noise, blur, and resolution degradation, ensuring diverse yet realistic degradations.
We adopt the BBDM~\cite{li2023bbdm} to learn a direct mapping between $\mathbf{x}_0$ and $\mathbf{y}$. Specifically, given an intermediate state $\mathbf{x}_t$ sampled from the forward bridge process, the model predicts the reverse mean using a UNet parameterization:
\begin{equation}
\mu_\theta(\mathbf{x}_t, t) \approx \tilde{\boldsymbol{\mu}}_t,
\end{equation}
and is optimized using an $\ell_1$ reconstruction loss:
\begin{equation}
\mathcal{L} = \mathbb{E}_{t,\mathbf{x}_0,\mathbf{y},\boldsymbol{\epsilon}} 
\left[
\left\|
m_t(\mathbf{y} - \mathbf{x}_0) + \sqrt{\delta_t}\boldsymbol{\epsilon} - \mu_\theta(\mathbf{x}_t, t)
\right\|_1
\right].
\end{equation}

To improve efficiency, images are encoded into a latent space using a pretrained VQGAN~\cite{esser2021vqgan}, and diffusion is performed at a reduced spatial resolution. The training pipeline validates dataset consistency, ensures one-to-one pairing between degraded and clean images, and leverages GPU acceleration for scalable optimization. 
By learning to reconstruct only bona fide iris textures, the model implicitly captures the underlying distribution of genuine samples, without exposure to attack data.

\subsection{Performance metrics}
To assess the detection performance of our model in Iris PAD, we adopted widely used evaluation metrics aligned with current SOTA benchmarks and ISO standards[13] in Presentation Attack Detection. These include:

\begin{itemize}

\item \textbf{APCER (Attack Presentation Classification Error Rate):}  
Proportion of attack presentations (from the same presentation attack instrument (PAI) species) that are incorrectly classified as bona fide presentations.
\begin{equation}
\text{APCER} = \frac{\text{Incorrectly classified PAI samples}}{\text{Total PAI samples}}
\end{equation}

\item \textbf{BPCER (Bonafide Presentation Classification Error Rate):}  
Proportion of bona fide presentations that are incorrectly classified as presentation attacks.
\begin{equation}
\text{BPCER} = \frac{\text{Incorrectly classified bonafide samples}}{\text{Total bonafide samples}}
\end{equation}

\item \textbf{ACER (Average Classification Error Rate):}  
Average of APCER and BPCER. Note: This metric is deprecated in ISO/IEC 30107-3:2017 but is still commonly reported for comparison with prior work.
\begin{equation}
ACER = \frac{APCER + BPCER}{2}
\end{equation}

\end{itemize}

\subsection{Inference and Decision Rule}
\label{sec:inference}

During inference, each input image $\mathbf{I}$ either bona fide or attack is first transformed into a degraded version $\mathbf{x}_0 = \mathcal{C}(\mathbf{I})$, consistent with the training pipeline. The trained Diff-IrisPAD model then reconstructs an output $\hat{\mathbf{y}}$ through the reverse diffusion process.
To quantify reconstruction fidelity, ViT Reconstruction Score, are computed:



\begin{equation}
\text{label} =
\begin{cases}
\text{attack}, & \text{if } S > \tau, \\
\text{bonafide}, & \text{otherwise}.
\end{cases}
\end{equation}

The complete scoring and threshold optimization pipeline, including multi-metric evaluation and weight selection, follows a validation-driven strategy.

\noindent\textbf{Threshold selection (zero-attack-knowledge (ZAK), unsupervised):} 
$\tau$ is set as the 90th percentile of bona-fide validation reconstruction-error scores ($\approx$10\% BPCER), with no use of attack data. 
The threshold is estimated solely from bona fide validation samples and is applied unchanged across all attack types, strictly adhering to the ZAK protocol.
The same $\tau$ is applied to 50{,}612 attacks and 12{,}926 bona-fide test images, and APCER, BPCER, ACER, and EER are computed per attack. Training and thresholding rely solely on bona-fide data, with attacks seen only at test time.

\section{Experiments}
We evaluate the proposed Diff-IrisPAD framework on a large-scale multi-sensor iris PAD benchmark under diverse attack conditions. The experiments are designed to assess both detection performance and generalization to unseen attacks.

\subsection{Datasets}

\begin{figure}[ht]
\centering
\includegraphics[width=0.7\linewidth]{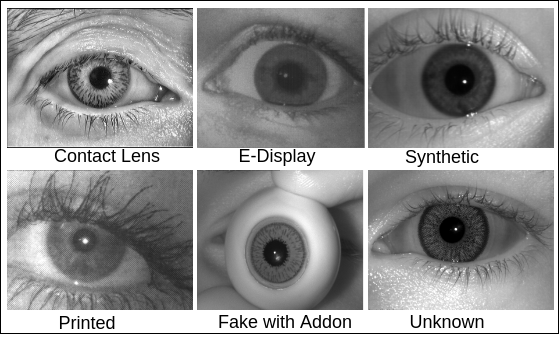}
\caption{Sample images illustrating different iris presentation attacks, including contact lens, display, synthetic and other}
\label{fig:DATASET}
\end{figure}

We evaluate the proposed method on a comprehensive iris PAD benchmark comprising both bona fide and diverse attack samples. The dataset includes multiple attack types such as artefacts, contact lenses, electronic displays, synthetic irises, printed attacks, additive fakes, and post-mortem samples. Representative examples of each attack category are shown in Figure \ref{fig:DATASET}.

\textbf{Clarkson (2013, 2015, 2017, 2023) - } This composite dataset, collected at Clarkson University using the LG IrisAccess EOU2200 sensor under NIR illumination, includes bona fide images along with textured contact lens and print attacks generated from both NIR and visible images. It contains over 20,000 images at a resolution of 600$\times$400.

\textbf{NDCLD (2013, 2015, 2017)}~\cite{doyle2014notre,doyle2015robust,yambay2017livdet} -  This dataset combines three subsets focused on contact lens attacks, including soft and textured lenses alongside bona fide samples. It contains over 10,000 NIR images captured at 640$\times$480 resolution using IrisAccess LG4000 and IrisGuard AD100 sensors.

\textbf{IIIT\_WVU~\cite{iab_iris} - } Designed for cross-dataset PAD evaluation, this dataset includes 6,250 training images and 4,209 test images captured using different sensors and environments. It contains bona fide samples, textured contact lenses, and print-based attacks, including combined attack scenarios.

\textbf{Warsaw (V1.1, V2, V3)} ~\cite{trokielewicz2016human, trokielewicz2018iris,trokielewicz2020post} - This dataset captures post-mortem iris images under both NIR and visible spectra using IriShield M2120U and Olympus TG-3 sensors. It includes over 2,800 NIR and 3,600 RGB images from 90+ subjects, enabling temporal analysis of iris degradation.

\textbf{IIITD Combined Spoofing}~\cite{kohli2016detecting, yadav2014unraveling, gupta2014iris} - A large-scale benchmark aggregating multiple public datasets, containing over 20,000 images with diverse attack types including contact lenses, print-recapture, and synthetic iris generation. Images are captured at 640$\times$480 resolution across multiple sensors.

\textbf{GUC\_VSIA\_CV} ~\cite{raghavendra2014presentation} - This dataset contains RGB iris images from 52 subjects, including 4,327 bona fide samples and various presentation attacks such as print and display-based spoofing, captured under indoor conditions using DSLR and light-field cameras.

\textbf{BTAS\_MIR2016}~\cite{zhang2016btas} -  This dataset includes multi-sensor iris images with both bona fide and spoof samples, designed to evaluate PAD performance under realistic acquisition conditions.
For datasets without predefined protocols, fixed training, validation, and test splits were established prior to model development to ensure a fair and reproducible evaluation.
\subsection{Preprocessing}
\label{sec:data_preprocessing}

All 125,854 iris images are preprocessed using a unified pipeline and resized to a fixed resolution of $256{\times}256$ RGB format for training, validation, and testing. 

\noindent\textbf{Iris localisation:}
To handle variability in sensor, resolution, and framing, we localise the iris region using the Circular Hough Transform~\cite{illingworth1988hough}. Each image is first downscaled to a 512-pixel thumbnail for efficient processing, followed by grayscale conversion, Gaussian smoothing ($5{\times}5$), and CLAHE-based contrast enhancement. Circle detection is then performed using \texttt{cv2.HoughCircles}, and the detected region is mapped back to the original resolution and cropped with a small radial margin. When detection fails, a centre-crop fallback is applied to ensure robustness. All crops are resized to $256{\times}256$ and converted to RGB.

\noindent\textbf{Degradation modelling:}
To facilitate paired learning, degraded inputs are generated from clean iris crops through a deterministic three-stage process comprising additive Gaussian noise, Gaussian blur, and resolution degradation via downsampling and upsampling. All parameters are deterministically seeded based on the image filename, ensuring consistent and reproducible transformations across training and inference.
\noindent\textbf{Data split:}
The dataset is divided into training and validation sets using only bona fide samples, ensuring no attack images are seen during training. The degradation process introduces substantial variation, enabling robust learning of clean iris representations.
\vspace{-0.1in}
\begin{table}[t]
  \centering
  \scriptsize
\setlength{\tabcolsep}{2pt}
  \caption{Diff-IrisPAD is trained using only bona-fide images, without attack data for training or thresholding. Evaluation uses a merged ZAK attack pool (50,612 samples) against 12,926 bona-fide test images, while 5,276 validation bona-fide images are reserved solely for threshold calibration.}
  \label{tab:dataset_splits}
  \small
  \begin{tabular}{lclll}
    \toprule
    Dataset          & Train & Val  & Test & Total     \\
    \midrule
    Bonafide        & 28056  & 5276  & 12926 & 46258 \\ 
    Artifact        & -      & 704   & 1643    & 2347 \\ 
    Contact Lens    & -      & 2840  & 6304    & 9144 \\
    E-display(ED)   & -      & 2369  & 5546    & 7915 \\ 
    Fake with Add-On & -     & 39   & 67      & 106 \\
    Generated        & -     & 2986   & 4978    & 7964 \\
    Post-Mortem     & -      & 2196   & 3293    & 5489 \\ 
    Print \& ED     & -      & 2220   & 5180    & 7400 \\ 
    Printed         & -      & 2750   & 7497    & 10247 \\
    \bottomrule
    \textbf{Total} & - & 16104 & 34508 & \textbf{50612} \\
    \bottomrule
  \end{tabular}
\end{table}

\subsection{Bona Fide Pair Construction}
\label{sec:pair_construction}

To enable unsupervised learning, we construct paired training data using only bona fide iris images. Each clean image is paired with a synthetically degraded version generated through a deterministic degradation pipeline, as shown in Figure.~\ref{fig:Diff-iris-PAD}.
The degradation process consists of three sequential steps: (i) Gaussian noise injection, (ii) Gaussian blur, and (iii) resolution degradation via downsampling and upsampling. The degradation parameters are randomly sampled within predefined ranges but remain deterministic for each image using a filename-based seed, ensuring reproducibility.
Formally, for each clean image \(\mathbf{y}\), a degraded counterpart \(\mathbf{x}_0\) is generated as:
\begin{equation}
\mathbf{x}_0 = \mathcal{C}(\mathbf{y}),
\end{equation}
where \(\mathcal{C}(\cdot)\) denotes the degradation function. The resulting dataset consists of aligned pairs \((\mathbf{x}_0, \mathbf{y})\), where \(\mathbf{x}_0\) serves as the input and \(\mathbf{y}\) as the reconstruction target.
Importantly, only bona fide images are used during training, and spoof samples are explicitly excluded to enforce an unsupervised learning paradigm.

\subsection{Training and Evaluation Dataset Construction}
\label{sec:dataset_construction}

For training, we use the constructed bona fide pairs, where corrupted images act as inputs and clean images as targets. The dataset is organized into aligned directories to ensure one-to-one correspondence between input-output pairs during training.
During evaluation, both bona fide and attack samples are considered. Clean images are first loaded and then corrupted on-the-fly using the same deterministic pipeline employed during training. This ensures consistency between training and testing distributions while enabling fair evaluation of PAD performance.
Each evaluation sample is associated with a label (bonafide or attack) and an attack type, allowing detailed performance analysis across different attack categories. The evaluation dataset is organized in a structured format with corresponding labels, ensuring compatibility with standard PAD evaluation protocols.




\subsection{Implementation Details}
\label{sec:implementation}

Diff-IrisPAD is implemented in PyTorch using the LBBDM-f4 configuration. Images are resized to $256 \times 256$ and encoded into a $64 \times 64$ latent space using a pretrained VQGAN. The model is trained on bona fide pairs with GPU acceleration. Evaluation follows a standard pipeline with validation-based threshold optimization. 

\subsection{Comparison with State-of-the-Art}

Table\ref{tab:diffusion_comparison} compares Diff-IrisPAD with diffusion-based iris PAD methods under a consistent unsupervised setting. The baseline AnoDDPM exhibits high error rates and unstable performance across attack types, with particularly large ACER values for Contact Lens (48.45), Generated (48.95), and Print \& E-display (54.07), resulting in a high overall ACER of 49.17.
In contrast, Diff-IrisPAD consistently achieves lower error rates across all attack categories. Notably, it obtains the lowest ACER for Artifact (6.08), Fake with Add-On (7.01), Post-Mortem (7.82), and Print \& E-display (25.41), while also reducing error on challenging attacks such as E-display (25.33) and Generated (46.96). 
Overall, Diff-IrisPAD achieves a substantially lower average ACER of \textbf{30.52}, significantly outperforming AnoDDPM. This improvement demonstrates that modeling bona fide iris distributions via Brownian Bridge Diffusion enables more stable and accurate detection compared to generic diffusion-based reconstruction approaches.

To ensure a fair comparison under the ZAK protocol, we compare Diff-IrisPAD with AnoDDPM, the closest diffusion-based unsupervised baseline, and f-AnoGAN, a representative GAN-based anomaly detection method. This comparison evaluates whether the observed improvements arise from the proposed DiffIrisPAD model formulation rather than reconstruction-based anomaly detection alone.
\vspace{-0.1in}
\begin{table}[!htbp]
\centering
\scriptsize
\setlength{\tabcolsep}{3pt}
\caption{
\textbf{Comparison of diffusion-based iris PAD methods.}
Threshold (unsupervised): $\tau$ = 90th percentile of bona-fide validation scores ($\approx$10\% BPCER), with no use of attack data.
BPCER consistency: A fixed $\tau$ on a shared bona-fide test set (12,926) yields identical BPCER across attacks; only APCER varies.
Test pool: 50,612 attacks (val+test) vs 12,926 bona-fide. Diff-IrisPAD uses $T{=}100$; AnoDDPM uses ViT-B/16 scoring. \textbf{Bold} = best ACER per attack.
}
\label{tab:diffusion_comparison}
\begin{tabular}{l l c c c c c}
\toprule
\textbf{Model} & \textbf{Attack} & APCER & BPCER & ACER & EER \\
\midrule
    \multirow{9}{*}{\shortstack[l]{f-AnoGAN\\($p{=}90$,\\$\tau{=}0.977$)\cite{schlegl2019f}}}

&Artifact      & 99.06 & 9.55 & 54.31 & 60.00 \\
&Contact Lens  & 98.34 & 9.55 & 53.95 & 56.48 \\
&E-display     & 87.96 & 9.55 & 48.76 & 41.20 \\
&Fake W/AO     & 99.06 & 9.55 & 54.31 & 68.18 \\
&Generated     & 98.93 & 9.55 & 54.24 & 60.63 \\
&Post-Mortem   & 93.39 & 9.55 & 51.47 & 46.87 \\
&Print \& ED   & 84.43 & 9.55 & 46.99 & 40.05 \\
&Printed       & 91.77 & 9.55 & 50.66 & 52.54 \\
\midrule

    \multirow{9}{*}{\shortstack[l]{AnoDDPM\\DDIM\\Simplex\\($p{=}90$,\\$\tau{=}0.859$)\cite{ho2020denoising}}}


& Artifact      &  98.17 &  9.21 & 53.69 &  50.83 \\
& Contact Lens  &  87.70 &  9.21 &  48.45 & 45.22 \\
& E-display     &  98.65 &  9.21 & 53.93 &  75.27 \\
& Fake W/AO     &  78.30 &  9.21 & 43.75 &  36.82 \\
& Generated     &  88.69 &  9.21 & 48.95 &  44.99 \\
& Post-Mortem   &  77.21 &  9.21 & 43.21 &  26.85 \\
& Print \& ED   &  98.93 &  9.21 & 54.07 &  63.72 \\
& Printed       &  80.77 &  9.21 & 44.99 &  47.67 \\
\midrule


\multirow{9}{*}{\shortstack[l]{\textbf{Diff-IrisPAD}\\(Ours,\\$T{=}100$)\\($p{=}90$,\\$\tau{=}0.164$)}}
& Artifact          &   0.98 & 11.19 &  \textbf{6.08} &  \textbf{5.34} \\
& Contact Lens      &  82.33 & 11.19 & \textbf{46.76} & \textbf{40.99} \\
& E-display         &  39.48 & 11.19 & \textbf{25.33} & \textbf{21.78} \\
& Fake W/AO         &   2.83 & 11.19 &  \textbf{7.01} &  \textbf{6.53} \\
& Generated         &  82.73 & 11.19 & \textbf{46.96} & \textbf{37.54} \\
& Post-Mortem       &   4.45 & 11.19 &  \textbf{7.82} &  \textbf{8.28} \\
& Print \& ED       &  39.64 & 11.19 & \textbf{25.41} & \textbf{21.57} \\
& Printed           &  46.67 & 11.19 & \textbf{28.93} & \textbf{23.42} \\
& \textbf{OverAll}      &  49.84 & 11.19 & \textbf{30.52} & \textbf{27.85} \\
\bottomrule
\end{tabular}
\end{table}
\vspace{-0.1in}
Table \ref{tab:diffusion_comparison} compares Diff-IrisPAD with representative unsupervised PAD methods under the same ZAK evaluation protocol.
\begin{table}[!htbp]
\centering
\scriptsize
\setlength{\tabcolsep}{3pt}
\caption{Leave-One-Attack-Out (LOO) evaluation. The model is trained excluding one attack type and tested on the unseen attack.}
\label{tab:loo}
\begin{tabular}{l l c c c c}
\toprule
\textbf{Model} & \textbf{Attack} & $\tau$ & APCER & BPCER & ACER \\
\midrule

    \multirow{9}{*}{\shortstack[l]{ResNet50 \cite{koonce2021resnet}}}
    &Artifact	    &0.45 &0.776	&0.021	&0.399	\\
    &CL	            &0.275 &0.990	&0.005	&0.497	 \\	
    &E-display   	&0.45 &0.319	&0.017	&0.168		\\
    &Fake W/AO  	&0.5 &0.134	&0.015	&0.074		\\
    &Generated  	&0.5 &0.458	&0.014	&0.236		\\
    &PostMortem	    &0.575 &0.353	&0.014	&0.183	\\	
    &Print \& ED	&0.475 &0.005	&0.014	&0.010	\\	
    &Printed	    &0.425 &0.1837	&0.016	&0.099	\\	
     \midrule  
   
   \multirow{9}{*}{\shortstack[l]{ViT-B \cite{dosovitskiy2020image}}}
    &Artifact	&0.7	&0.731	&0.030	&0.380 \\ 
    &CL    	    &0.85	&0.966	&0.009	&0.488 \\ 
    &E-display	&0.525	&0.296	&0.032	&0.164 \\ 
    &Fake W/AO	&0.475	&0	    &0.039	&0.019 \\
    &Generated	&0.5	&0.466	&0.035	&0.251  \\
    &PostMortem	&0.5	&0.517	&0.0311	&0.274  \\
    &Print \& ED &0.5	&0.073	&0.049	&0.061 \\
    &Printed	&0.475	&0.441	&0.032	&0.237 \\
     \midrule 
	
    \multirow{9}{*}{\shortstack[l]{MaxViT \cite{tu2022maxvit}}}
    &Artifact	&0.5	&0.869	&0.025	&0.447 \\
    &CL	        &0.375	&0.963	&0.008	&0.486 \\
    &E-display	&0.4	&0.122	&0.019	&0.070 \\
    &Fake W/AO	&0.425	&0	    &0.025	&0.012 \\
    &Generated	&0.475	&0.521	&0.018	&0.270 \\
    &PostMortem	&0.5 	&0.440	&0.025	&0.233 \\
    &Print \& ED &0.425	&0.009	&0.023	&0.016 \\
    &Printed	&0.45	&0.331	&0.019	&0.175 \\

     \midrule 
    \multirow{9}{*}{\shortstack[l]{DINOv1 \\ ResNet50\\(Backbone) \cite{caron2021emerging}}}
    
    &Artifact	    &0.425	&0.858	&0.019	&0.438  \\
    &CL	            &0.425	&0.971	&0.005	&0.488  \\
    &E-display	    &0.55	&0.405	&0.017	&0.211 \\
    &Fake W/AO	    &0.525	&0	    &0.015	&0.007 \\
    &Generated      &0.475	&0.503	&0.017	&0.260 \\
    &PostMortem  	&0.475	&0.616	&0.020	&0.318 \\
    &Print \& ED	&0.35	&0.007	&0.018	&0.013 \\
    &Printed	    &0.65	&0.146	&0.008	&0.077 \\
    
     \midrule  
	
    \multirow{9}{*}{\shortstack[l]{DINOv2 + \\ ViTs14\\(Backbone)\cite{oquab2023dinov2}}}
    
    &Artifact	    &0.6	  &0.720	&0.017	&0.369 \\
    &CL	            &0.275	  &0.973	&0.009	&0.491 \\
    &E-display      &0.5	  &0.100	&0.032	&0.066 \\
    &Fake W/AO  	&0.525	  &0	    &0.011	&0.005 \\
    &Generated	    &0.5	  &0.599	&0.020	&0.309 \\
    &PostMortem	    &0.4	  &0.463	&0.033	&0.248 \\
    &Print \& ED	&0.5	  &0.000	&0.023	&0.012 \\
    &Printed	    &0.575	  & 0.174	&0.012	&0.093 \\
\bottomrule
\end{tabular}
\end{table}
Contact Lens and Generated attacks remain challenging due to their close resemblance to bona fide iris textures. Similar performance trends are also observed in supervised methods under unseen attack settings. Despite relying solely on bona fide training samples, Diff-IrisPAD achieves comparable performance, demonstrating the effectiveness of attack-independent bona fide modeling.
\subsection{Generalization to Unseen Attacks}
Table \ref{tab:loo} illustrates the challenges of unseen attacks and the limitations of attack-dependent learning, thereby motivating the proposed attack-independent bona fide modeling approach. It presents the leave-one-attack-out (LOO) evaluation, where models are trained by excluding one attack type and evaluated on the corresponding unseen attack. 
Supervised models, including CNN (ResNet50), transformer (ViT-B, MaxViT), and foundation models (DINOv1, DINOv2), achieve strong performance on certain attack types but exhibit significant variability across unseen conditions. In particular, high ACER values are observed for challenging attacks such as Contact Lens (e.g., 0.497 for ResNet50, 0.488 for DINOv1) and Artifact (e.g., 0.399 for ResNet50), indicating limited robustness when attack characteristics differ from training data.
Transformer and foundation models show improved performance on some attacks (e.g., DINOv2 achieves low ACER for Printed and Fake with Add-On), but still lack consistent generalization across all attack types.
These results highlight that supervised methods, while effective on known attacks, are sensitive to distribution shifts. In contrast, Diff-IrisPAD, trained solely on bona fide samples, avoids reliance on attack-specific supervision and is inherently better suited for open-set scenarios. By modeling the distribution of genuine iris textures, it provides a more robust and stable framework for detecting unseen presentation attacks.

\begin{figure}[ht]
\centering
\includegraphics[width=\columnwidth]{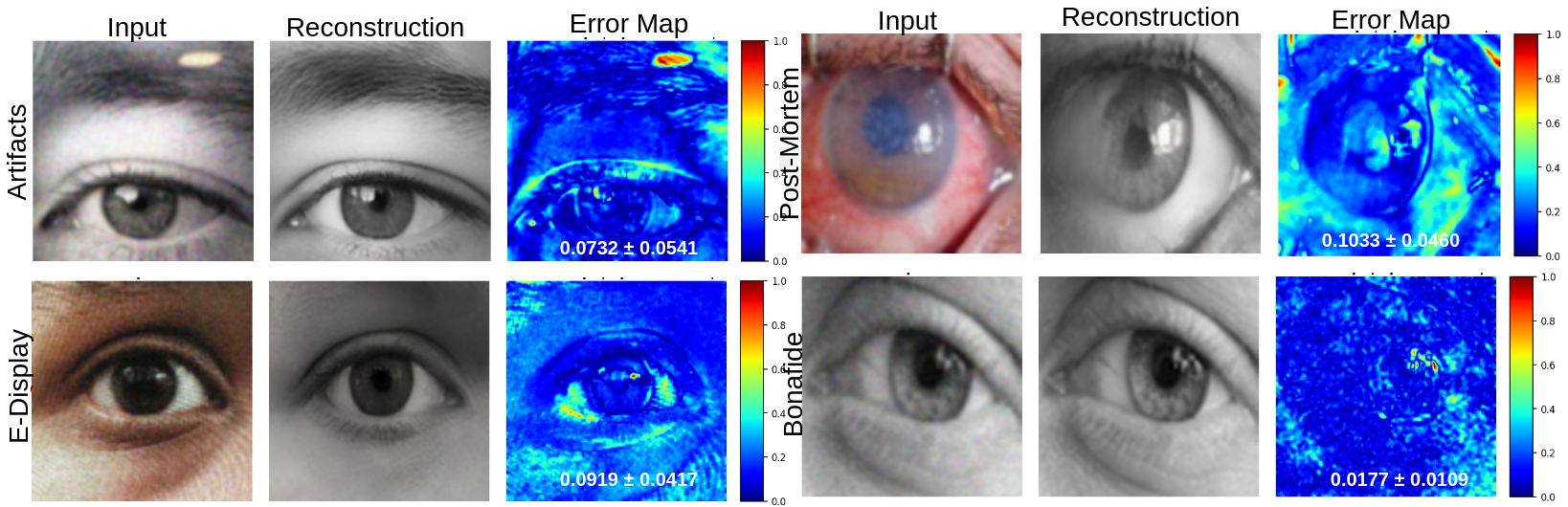}
\caption{Reconstruction and error analysis. Attack samples show higher reconstruction error, while bona-fide images exhibit low error, enabling clear separation.}
\label{fig:recon_analysis}
\end{figure}

\noindent\textbf{Reconstruction and Error Analysis.}
Figure~\ref{fig:recon_analysis} shows qualitative reconstruction results and corresponding error maps across different samples. Bona fide inputs are reconstructed accurately with low error responses, while attack samples exhibit noticeable reconstruction discrepancies and higher error regions. These high-error regions correspond to deviations from the learned bona fide distribution, particularly around iris texture and boundary areas. This behavior confirms that Diff-IrisPAD effectively captures genuine iris characteristics while failing to reconstruct attack-specific artifacts, enabling reliable separation between bona fide and attack samples.

\section{DET Curve Analysis}
\label{sec:det_curves_sup}
We present Detection Error Tradeoff (DET) curves for multiple presentation attack types across CNN, transformer, foundation, diffusion, and the proposed Diff-IrisPAD models. For each attack, DET curves compare models under a consistent evaluation protocol, providing insight into the APCER BPCER trade-off beyond the threshold based metrics in Table~\ref{tab:loo} \& DET curve.
(see supplementary material).
\vspace{-0.1in}
\subsection{Ablation Study}
\label{sec:ablation}

\noindent\textbf{Effect of Denoising Steps:}
Table~\ref{tab:ablation_study2} evaluates the impact of DDIM denoising steps ($T$) on PAD performance. Increasing $T$ improves reconstruction quality and stabilizes detection across attack types. 
Low step counts ($T{=}10$) offer fast inference (84.4 ms/img) but result in higher ACER due to limited reconstruction fidelity. Increasing to $T{=}50$ significantly improves stability across challenging attacks such as E-display and Generated samples. 
The best performance is achieved at $T{=}100$, yielding the lowest overall ACER of \textbf{30.52} with consistent APCER/BPCER across attacks, indicating an optimal balance between accuracy and efficiency. While further increasing to $T{=}200$ provides marginal gains in certain cases (e.g., Artifact), it leads to higher computational cost (671.9 ms/img) without improving overall performance. 
Overall, moderate denoising steps ($T{=}100$) are sufficient to capture discriminative iris structures while maintaining computational efficiency.
\vspace{-0.1in}
\begin{table}[!htbp]
\centering
\scriptsize
\setlength{\tabcolsep}{3pt}
\caption{\textbf{Ablation on denoising steps ($T$).}
LBBDM-f4 with ViT scoring; $\tau$ tuned on validation data. Results averaged across attacks (lower is better).}
\label{tab:ablation_study2}
\begin{tabular}{l l c c c c c}
\toprule
\textbf{Steps} & \textbf{Attack} & APCER & BPCER & ACER & EER & ms/img \\
\midrule


\multirow{9}{*}{\shortstack[l]{10\\$\tau{=}0.201$}}
& Artifact          &  1.87 &  9.68 &           5.78 &  6.23 & \multirow{9}{*}{84.4} \\
& CL                & 85.71 &  9.68 &          47.69 & 41.58 & \\
& E-display         & 53.24 &  9.68 &          31.46 & 26.73 & \\
& Fake W/AO         &  6.60 &  9.68 &           8.14 &  7.55 & \\
& Generated         & 88.16 &  9.68 &          48.92 & 42.35 & \\
& Post-Mortem       & 11.81 &  9.68 &          10.74 & 10.58 & \\
& Print \& ED       & 53.64 &  9.68 &          31.66 & 26.70 & \\
& Printed           & 56.42 &  9.68 &          33.05 & 27.07 & \\
& Overall           & 58.33 &  9.68 &          34.00 & 30.98 & \\
\midrule

\multirow{9}{*}{\shortstack[l]{50\\$\tau{=}0.170$}}
& Artifact          &  0.89 & 10.66 &           5.78 &  5.18 & \multirow{9}{*}{208.6} \\
& Contact Lens      & 83.73 & 10.66 &          47.19 & 40.95 & \\
& E-display         & 41.07 & 10.66 &          25.87 & 21.95 & \\
& Fake W/AO         &  2.83 & 10.66 &           6.75 &  7.55 & \\
& Generated         & 83.86 & 10.66 &          47.26 & 37.91 & \\
& Post-Mortem       &  5.78 & 10.66 &           8.22 &  8.50 & \\
& Print \& ED       & 40.51 & 10.66 &          25.59 & 21.53 & \\
& Printed           & 48.19 & 10.66 &          29.43 & 23.35 & \\
& Overall           & 51.10 & 10.66 &          30.88 & 27.73 & \\
\midrule

\multirow{9}{*}{\shortstack[l]{200\\$\tau{=}0.167$}}
& Artifact          &  0.85 & 10.97 &           5.91 &  5.30 & \multirow{9}{*}{671.9} \\
& Contact Lens      & 83.14 & 10.97 &          47.05 & 40.88 & \\
& E-display         & 40.48 & 10.97 &          25.73 & 22.09 & \\
& Fake W/AO         &  2.83 & 10.97 &           6.90 &  6.44 & \\
& Generated         & 82.92 & 10.97 &          46.95 & 37.69 & \\
& Post-Mortem       &  5.28 & 10.97 &           8.13 &  8.38 & \\
& Print \& ED       & 40.51 & 10.97 &          25.74 & 21.82 & \\
& Printed           & 48.16 & 10.97 &          29.57 & 23.95 & \\
& Overall           & 50.69 & 10.97 &          30.83 & 27.78 & \\
\bottomrule
\end{tabular}
\end{table}
\vspace{-0.1in}

\noindent\textbf{Limitations.}
Diff-IrisPAD depends on effective degradation modeling; mismatched synthetic degradations can reduce reconstruction quality. It may also struggle with highly realistic attacks that closely resemble bona fide iris textures. 

\section{Conclusion}

This paper presented Diff-IrisPAD, an unsupervised iris PAD framework based on BBDM model. By learning only the bona fide iris manifold, it eliminates the need for attack samples and achieves an ACER of 30.52, outperforming AnoDDPM \& f-AnoGAN. These results demonstrate that attack independent bona fide modeling improves generalization to unseen presentation attacks under the ZAK protocol. Future work will investigate adaptive degradation and cross-domain generalization.



{\small
\bibliographystyle{ieee}
\bibliography{egbib}
}

\end{document}